\documentclass[10pt,twocolumn,letterpaper]{article}

\usepackage{iccv}
\usepackage{times}
\usepackage{epsfig}
\usepackage{graphicx}
\usepackage{amsmath}
\usepackage{amssymb}

\usepackage{latexsym}

\usepackage{microtype}
\usepackage{inconsolata}
\usepackage{latexsym}
\usepackage{xcolor}
\usepackage{makecell}
\usepackage{latexsym}
\usepackage{hyperref}
\usepackage{inconsolata}
\usepackage{url}
\usepackage{makecell}
\usepackage{multirow}
\usepackage{graphicx}
\usepackage{caption}
\usepackage{subcaption}
\usepackage{amsmath}
\usepackage{bbm}
\usepackage{booktabs}
\usepackage{algorithmic}
\usepackage{amssymb}
\usepackage{bm}
\usepackage[ruled,vlined,linesnumbered]{algorithm2e}
\usepackage{enumitem}
\usepackage{pifont}
\usepackage{makecell}
\newcommand{\xmark}{\ding{55}}%
\newcommand{\cmark}{\ding{51}}%
\usepackage{color}

\usepackage{indentfirst}

\iccvfinalcopy 

\ificcvfinal\fi

\begin{document}

\title{ \textit{DocumentCLIP}: Linking Figures and Main Body Text in Reflowed Documents}

\author{Fuxiao Liu\\
University of Maryland, College Park\\
{fl3es@umd.edu}
\and
Hao Tan\\
Adobe Research\\
{hatan@adobe.com}
\and
Chris Tensmeyer\\
Adobe Research\\
{tensmeye@adobe.com}
}

\maketitle
\ificcvfinal\thispagestyle{empty}\fi

\begin{abstract}
Vision-language pretraining models have achieved great success in supporting multimedia applications by understanding the alignments between images and text. While existing vision-language pretraining models primarily focus on understanding single image associated with a single piece of text, they often ignore the alignment at the intra-document level, consisting of multiple sentences with multiple images. In this work, we propose \textit{DocumentCLIP}, a salience-aware contrastive learning framework to enforce vision-language pretraining models to comprehend the interaction between images and longer text within documents. Our model is beneficial for the real-world multimodal document understanding like news article, magazines, product descriptions, which contain linguistically and visually richer content. To the best of our knowledge, we are the first to explore multimodal intra-document links by contrastive learning. In addition, we collect a large Wikipedia dataset for pretraining, which provides various topics and structures. Experiments show \textit{DocumentCLIP} not only outperforms the state-of-the-art baselines in the supervised setting, but also achieves the best zero-shot performance in the wild after human evaluation. Our code is available at \hyperlink{https://github.com/FuxiaoLiu/DocumentCLIP}{https://github.com/FuxiaoLiu/DocumentCLIP}.
\end{abstract}

\section{Introduction}
Images and text act as natural complements on the modern web. A number of works \cite{radford2021learning, li2019visualbert, lin2014microsoft, li2022blip, alayrac2022flamingo, li2023blip, liu2023covid, zhang2021vinvl} have been proposed to interpreting an image with the corresponding short text such as a caption or question. However, real-world media such as news articles, Wikipedia pages, magazines, product descriptions consist of multiple sentences with multiple images. Algorithms that identify document-internal connections between specific images and specific unit of text could have the long-term promise. For example, alt-text for vision-impaired users could be produced automatically via intra-document retrieval. Additionally, by designing a user interface hinting the readers that there is a figure associated with a part of the text upon users's action, it can not only help users to understand the whole document, but also make it easy and comfortable to read on smart phones, where the font and button is small.

\begin{figure*}[t!]
    \centering
      \includegraphics[width=1\textwidth]{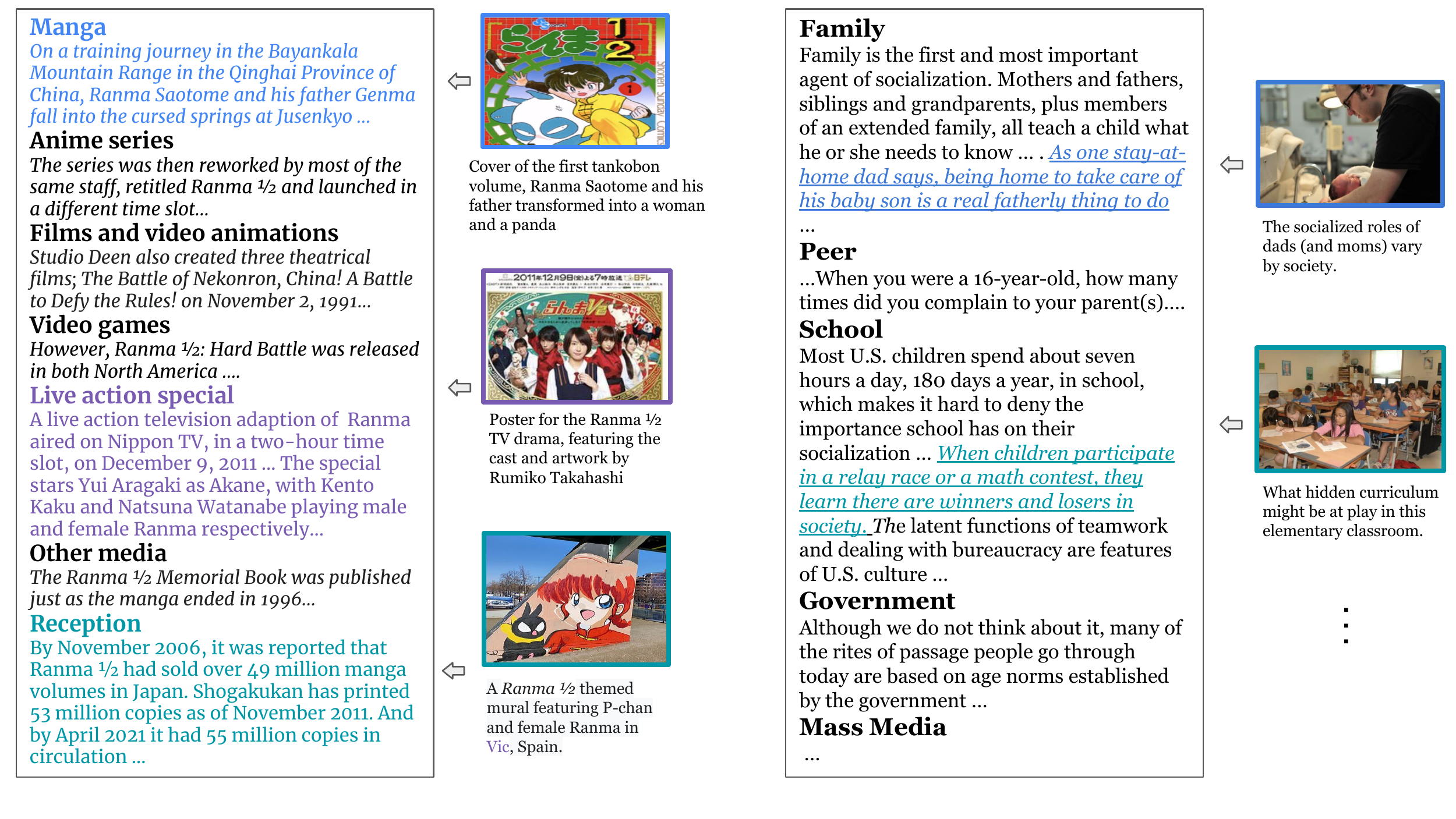}
    \caption{Illustration of our task. Given an article, our model extracts the most relevant text to the image/caption pair. (Left): The document is from our pretraining \textit{Wikipedia dataset}, which has the section to image/caption pair links as the groundtruth label. We leverage this weak supervision to explore multimodal intra-document interaction by contrastive learning. (Right): An example is from \textit{Open Textbook dataset} in the wild. The highlights are the most relevant sentences extracted by our model.}
    \vspace{-0.2in}
    \label{fig:lead}
\end{figure*}

This intra-document setting introduces several challenges compared to conventional cross-model retrieval task with a single sentence and caption \cite{muraoka2020image,liu2020upgrading, nag2021sandi,zhang2021news}. More specifically, since the images are to be incorporated in the same article, their contents should be consistent according to the theme of the article, thus making disambiguation more difficult than in the usual one-image/one-sentence case (Figure~\ref{fig:lead}). Additionally, a sentence in the longer document may correspond to multiple images or no related images. Another challenge is that it requires considering longer texts with linguistically richer content and the relations between them, like the structure and layout information. The previous relevant works \cite{radford2021learning, li2019visualbert, yao2021filip, li2021align, li2020oscar, yu2022coca} trained to align short literal text descriptions of images with the image failed to encode the structure features. Document structure analysis models \cite{huang2022layoutlmv3, gu2021unidoc, xu2020layoutlmv2} process pages one by one, but the figures may relate to text located on other pages.

\begin{figure*}[t!]
    \centering
      \includegraphics[width=1\textwidth]{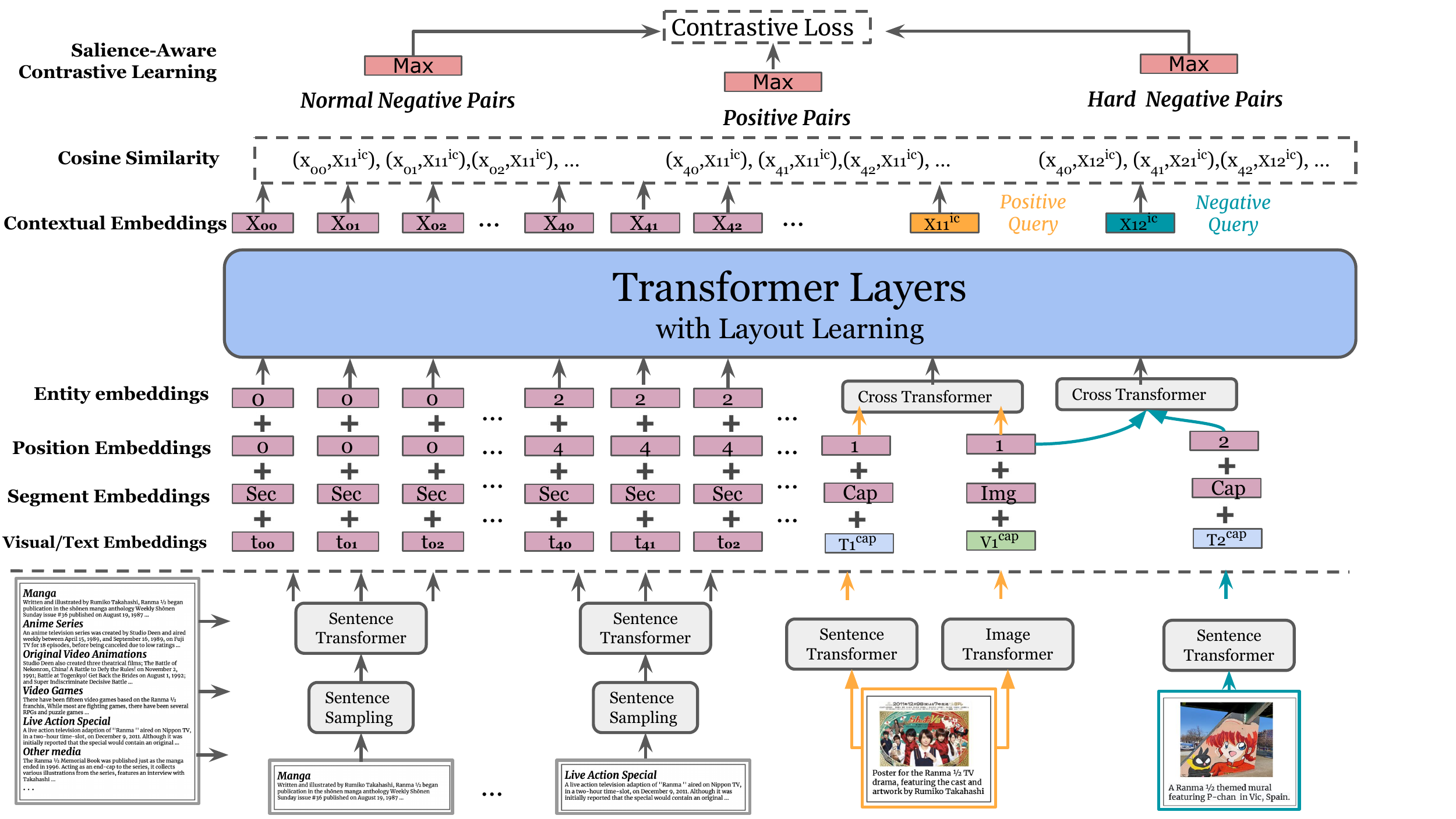}
    \caption{Overview of \textit{DocumentCLIP} in the training phrase. In this example, \texttt{$t_{00}, t_{01}, t_{02}$} are the Top 3 sentences selected by the \textit{Salient Sentence Extraction} from the image/caption pair in yellow. $x^{ic}_{12}$ is the negative query from the image/caption pair in green to generated hard negative pairs. $x^{ic}_{11}$ is the positive query to construct positive pairs and normal negative pairs. }
    \vspace{-0.2in}
    \label{fig:model}
\end{figure*}

To address the aforementioned challenges, we propose \textit{DocumentCLIP}, a simple yet effective pretraining learning method of intra-document vision and language understanding tasks. Inspired by the BERT model \cite{liu2019roberta}, where input textual information is mainly represented by text embeddings and position embeddings, \textit{DocumentCLIP} further adds two types of input embeddings: (1) an entity embedding that indicates which section sharing more common entities. This is because the images/caption pairs can be the literal visualization of entities mentioned in document sentences. (2) a section position embedding that denotes the relative position within a document to encode the layout and structure information. In addition, we introduce the salient sentence extraction strategy to extract the important information from the long section. To effectively transfer event knowledge across modalities, we employ early-fusion method to aggregate images and caption by \cite{li2020hero} before the layout transformer in Figure~\ref{fig:model}. Unlike the state-of-the-art vision-language pre- training model CLIP \cite{radford2021learning}, we optimize a salience-aware contrastive learning objective between intra-document components: images, captions, section text. In order to train robust representations capable of discriminating subtle differences between different sentences in the same document, we not only have normal negatives by replacing groundtruth section text with others from the same document, but also propose to generate hard negatives by manipulating the image and caption pairs by switching one of them.

In this work, we pretrain and evaluate \textit{DocumentCLIP} on the Wikipedia articles, which have natural correspondences between the text sections and image/caption pairs in its HTML sources and describe various topics. Also, our evaluations focus on zero-shot settings, since it is crucial to understand real-world documents with various topics and sources. Examples are shown in Figure~\ref{fig:lead}. Experiment results show that \textit{DocumentCLIP} significantly outperforms baselines and meet users' requirement in the wild\footnote{https://open.umn.edu/opentextbooks/}. The contributions of this paper are summarized as follows:
\begin{itemize}[nosep,leftmargin=*]
    \item We introduce a novel contrastive learning framework with layout information, multimodal interaction, novel section encoding strategy, salience-aware contrastive loss and hard negative samples.
    \item We collect a large visual article dataset from Wikipedia with various topics, containing 66k articles, 320k images/caption pairs.
    \item Our model outperforms baselines on the intra-document understanding task in both the supervised and zero-shot settings after human evaluation. 
\end{itemize}

\vspace{-0.12in}

\label{sec:intro}

\section{\textit{DocumentCLIP}}
\subsection{Problem Statement}
We assume that the structure of the document has been parsed, so that we can apriori identify figure images and their captions (if present), and segment the main body text into a sequence of sections, paragraphs, and sentences. The granularity of the associated text (section vs sentence) can be set based on the use case and/or training data available. In our prototype, we primarily used Wikipedia articles, whose source markup associates figure-captions with sections. Additionally, by leveraging this weak supervision with the section ground truth label, our model is often able to retrieve the most relevant sentence to the associated image and caption. 

We define relevance based on the following criteria. (1) The images are the literal visualization of the entities mentioned in the document. (2) Text explicitly refers to the images. (3) The images provide evidences for the text claims.

\subsection{Model Architecture}
We build a multi-modal Transformer architecture as the backbone of \textit{DocumentCLIP}, which takes text, images, and layout information as input to establish deep cross-modal interactions. We describe the strategies to encode sections of the document in \ref{subsection1}. In \ref{subsection2}, we present how \textit{DocumentCLIP} leans layout information and feature extraction. Finally, we discuss the objective functions and model inference in \ref{subsection3}. Detailed descriptions of the model are illustrated in Figure~\ref{fig:model}.

\subsubsection{Section Encoding Strategy}\label{subsection1}
Although BERT-like models become the state-of-the-art techniques on several challenging NLP tasks, they usually leverage only text information without image features. CLIP \cite{radford2021learning} models achieves great success in image-text retrieval task. However, it's only able to handle short text (less than 77), which is much shorter than the average length (195.5) of the sections within Wikipedia documents. Therefore, we experiment different sentence sampling strategies and propose the novel \textit{Salient Sentences Extraction} method to learn the section representation.

\textit{First Sentence.} It is evident that the first sentence plays an important role and always presents the central meaning of the sections. Therefore, we can embed the first sentence to represent the section.

\textit{Weighted Average.} Apart from the first sentence, the associated image can also provides evidences for other sentences in a section. Thus, we train a one-layer transformer to aggregate all the important sentence embeddings in a section. 

\textit{All Text Concat.} Compared with the \textit{Weighted Average} method, we regard the text in a section as one sentence and encode them together by a text transformer.

\textit{Salient Sentences.} Motivated by the success of transferring image-text pre-training knowledge, we first adopt CLIP \cite{radford2021learning} to compute a scaled dot product similarity score $Sim_{si}$ between each sentence in a section and images. Then we also do the operation $Sim_{sc}$ for sentences and captions. After that, we utilize the average value of the $Sim_{si}$ and $Sim_{sc}$ to represent the correlation between the sentence and image/caption pairs. Finally, we pick the top $K$ salient sentences in each section with the highest scores as the candidates. In Figure~\ref{fig:model}, $K$ is 3 for simplicity and candidates are generated from image/caption in yellow. 

\subsubsection{Document Layout Learning}\label{subsection2}
In this section, we extend to jointly model text, image and layout information in the \textit{DocumentCLIP} framework.

\textbf{Text Embedding.} Following the common practice, we use 
lower-cased byte pair encoding (BPE) \cite{shibata1999byte} to tokenize the sentences in documents. Then, we add \texttt{[CLS]} at the beginning of the sequence and \texttt{[SEP]} at the end. Extra \texttt{[PAD]} tokens are appended to the end so that the final sequence’s length is exactly the maximum sequence length \texttt{L}. We directly adopt CLIP for initialization to extend its ability. The final token embedding is the sum of token embeddings and positional embeddings. Formally, the sentence embedding is computed by the text transformer \cite{radford2019language} and output of \texttt{[CLS]} token which {$T^{sec}_j$}is utilized as sentence representations in documents. \texttt{$T^{sec}_{j}$} $\in$ \texttt{$[t_{j0}, ... , t_{jK}]$}, where $j$ $\in$ \texttt{(0 $\leq$ j < N)} means the section index in a document. \texttt{$[t_{j0}, ... , t_{jK}]$} are the top $K$ candidates selected by salient sentence extraction in the \texttt{j-th} section. Similarly, the caption representation is $T^{cap}_i$, where $i$ $\in$ \texttt{(0 $\leq$ i < M)} and \texttt{M} means the caption number in a document.

\begin{table*}[t]
\centering
\small
\begin{tabular}{c|c|c|c|c|c|c}
\toprule
 & KVQA & TQA& Visual Genome & GoodNews & Visual News & Wikipedia (Ours)\\
\midrule
\textit{Avg Unit Length} & $84.8$ & $920.8$ & $263.9$ & $451.0$ & $773.0$ & $195.5$ \\
\textit{Avg Doc Length} & $11.4$ & $106.0$ & $5.3$ & $18.0$ & $18.8$ & $3346.6$ \\
\textit{Avg Sent Length} & $11.4$ & $12.2$ & $5.3$ & $18.0$ & $18.8$ & $22.3$ \\
\textit{Avg Sent Num per Unit} & $1.0$ & $7.7$ & $1.0$ & $1.0$ & $1.0$ & $8.2$ \\
\textit{Avg Img Num per Doc} & $-$ & $3.0$ & $-$ & $1.8$ & $1.7$ & $4.8$ \\
\bottomrule
\end{tabular}
\caption{Statistics in terms of the document($\textit{Doc}$), unit and images($\textit{Img}$). $\textit{Doc}$ donates document and $\textit{Sent}$ donates sentence. $\textit{Unit}$ means the section in our dataset.}
\label{tab:dataset_summary}
\vspace{-0.2in}
\end{table*}
\textbf{Visual Embedding.} In order to obtain image embedding, vision transformer is adopted to extract N non-overlapping image patches and perform linear projection to map every patch into 1D token. With injection of positional embedding and extra [CLS] token, the sequence of tokens are fed into image transformer layers, to model the correlation of each patch, where each layer ls comprises of Multi-Head Self-Attention, layer normalization, and Multi-layer Perception. Then, output \texttt{[CLS]} token embedding is used to represent the image feature $V^{img}_k$, where $k$ $\in$ \texttt{(0 $\leq$ k < M)} and \texttt{M} means the image number in a document.

\textbf{Layout Embedding.} The layout embedding layer is to understand the global context of the document. Thus, The final text embedding of captions $\texttt{$l^{cap}_i$}$ is the sum of three embeddings. The sentence embedding $T^{cap}_i$ represents the caption itself. $\texttt{PosE(i)}$ is the positional embedding, where $i$ represents the caption index in the document. The segment embedding \texttt{SegE($T^{cap}_i$)=[C]} is used to distinguish different segments. For instance in Figure~\ref{fig:model}, the position index of the image is \texttt{1} since it's the second image in the document. Similarly, the final image embedding \texttt{$l^{img}_k$} is aggregated by $V^{img}_k$, \texttt{PosE}(k) and \texttt{SegE($V^{img}_k$)=[V]}. 
\begin{align}
\texttt{$l^{cap}_i$} &= T^{cap}_i + \texttt{PosE}(i) + \texttt{SegE}(T^{cap}_i)\\
\texttt{$l^{img}_k$} &= V^{img}_k + \texttt{PosE}(k) + \texttt{SegE}(V^{img}_k)\\
\texttt{$l^{sec}_{j}$} &= T^{sec}_j + \texttt{PosE}(j) + \texttt{SegE}(T^{sec}_j) + \texttt{EntE}(s_j)
\end{align}
Inspired by the fact that the complementary images/caption pairs can be the literal visualization of entities mentioned in document sentences, we calculate the common entity numbers between sections with captios and sort sections in the descending order. More formally, we introduce the entity embedding \texttt{EntE}$(s_j)$ for \texttt{$l^{sec}_{j}$}, where $s_j$ $\in$ \texttt{(0 $\leq$ i < N)}. If a section share most entities with a images/caption pair, $s_j$\texttt{=0}, while $s_j$\texttt{=N-1} if it shares the least entities. In Figure~\ref{fig:model}, given the positive query (image/caption in yellow), $s_0$\texttt{=0} and $s_4$\texttt{=2}.

\textbf{Image and Caption Fusion.} The intuitive way to select the related section is to compute the similarity score between section text with the image and caption separately. Then train a learnable coefficient to combine them. However, it ignores the potential correlation between the image and caption effectively. Therefore, we apply the Cross-modal Transformer \cite{li2020hero} to fuse the \texttt{$l^{cap}_i$} and \texttt{$l^{img}_k$} before the layout transformer. We also add the \texttt{[cls]} token in the first place of the input sequence. The outputs from Cross-modal Transformer is a sequence of contextualized embeddings. We use the output from the \texttt{[cls]} token as the unified representation of a image and caption pair.
\begin{align}
l^{ic}_{ki} &= \textit{Cross-Transformer}(\texttt{$l^{img}_k$}, \texttt{$l^{cap}_i$})
\end{align}

\textbf{Layout Transformer.} We concatenates section sentence embeddings \texttt{\{$l^{sec}_{0}$, $l^{sec}_{1}$, $...$, $l^{sec}_{N-1}$\}} and the image/caption pair \texttt{\{$l^{ic}_{00}$, $l^{ic}_{11}$, $...$, $l^{ic}_{(M-1)(M-1)}$\}} to a unified sequence $L$. In order to make the features more compact, we employ an FC layer with a ReLU activation before the layout Transformer \cite{liu2020visual}. Formally, we obtain the contextual multimodal embeddings $X$ after transformer layers with layout learning. More formally, $X=$ \texttt{\{$x^{sec}_{0}$, $...$ $x^{sec}_{N-1}$, $x^{ic}_{00}$, $...$, $x^{ic}_{(M-1)(M-1)}$}\}. \texttt{$x^{sec}_{j}$} $\in$ \texttt{$[x_{j0},..., x_{jK}]$} are the contextual embeddings of the Top $K$ sentences each section, where $j$ $\in$ \texttt{(0 $\leq$ j < N)} means the section index.
\begin{align}
X &= \textit{Layout-Transformer}(ReLU(FC(L)))
\end{align}

\subsubsection{Salience-Aware Contrastive Learning.} \label{subsection3}
In this section, we introduce the definition of positive/negative pairs for similarity learning and our salient-aware contrastive loss.. 

\textbf{Positive Pairs.} We define that a section $x^{sec}_j$ and image/caption pair $x^{ic}_{k,i}$ is positive if the section contains certain content that is relevant to both the image and caption in $x^{ic}_{k,i}$. For example in Figure~\ref{fig:model}, \texttt{$\beta^p=$} \texttt{\{$(x_{40}$,$x^{ic}_{11})$,$(x_{41}$, $x^{ic}_{11})$,$(x_{42}$,$x^{ic}_{11})$,...\}} are the positive pairs, where $x^{ic}_{11}$ is the positive query. \texttt{$[x_{40},x_{41},...,x_{4K}]$} are the Top $K$ sentences embeddings in the forth section.

\textbf{Negative Pairs.} Our negative pairs set $\beta^n$ contains two parts: normal negative pairs and hard negative pairs. In normal negative pairs, the groundtruth section is replaced by other irrelevant sections in the same document. In Figure~\ref{fig:model},  \texttt{\{$(x_{00}$, $x^{ic}_{11})$,$(x_{01}$, $x^{ic}_{11})$,$(x_{02}$,$x^{ic}_{11})$, $(x_{10}$,$x^{ic}_{11})$,$(x_{12}$,$x^{ic}_{11})$,$(x_{12}$,$x^{ic}_{11})$...\}} are the positive pairs, where $x^{ic}_{11}$ is the positive query. \texttt{$[x_{40}, x_{41}, ..., x_{4K}]$} are normal negative pairs. As for the hard negative pairs, either the caption items or the image items might be changed but not both of them. For example in Figure~\ref{fig:model}, $x^{ic}_{12}$ is the hard negative query and \texttt{\{$(x_{40}$, $x^{ic}_{12})$,$(x_{40}$, $x^{ic}_{21})$,$(x_{41}$,$x^{ic}_{12})$, $(x_{41}$,$x^{ic}_{21})$,...\}} are hard negative pairs. This is inspired by the fact that both the caption and image play important role to determine the relevant text from the document.

\textbf{Salience-Aware Contrastive Loss.} Motivated by the fact that the section label is used as the weak supervision, we propose to equip the model with the ability to extract most salient sentence from each section given the query $x^{ic}_{ki}$ and document context. Additionally, unrelated sentences in groundtruth section can introduce additional noise in the contrastive learning process. Formally, we measure the cosine similarity between query $x^{ic}_{ki}$ and \texttt{$[x_{j0},...,x_{jK}]$}. Then we select the maximum score to represent the correlation between the section and image/caption pair:
\begin{align} 
 & q = x^{ic}_{k,i},  s = x^{sec}_{j} \in \texttt{$[x_{j0},..., x_{jK}]$}\\
& S(q,s) = max\{cos(q,s)\}, (q,s) \in \beta^p  \\
& S^*(q,s) = max\{cos(q,s)\}, (q,s) \in \beta^n  \\
& \textit{$L$} = -\sum_{(q,s) \in D}^{}[\textit{log}(\frac{S(q,s)}{S(q,s)+\sum_{S^* \in \beta^n} S^*(q,s)}]
\end{align}
where $S(q,s)$ collects the maximum similarity scores of positive pairs and $S^*(q,s)$ includes the negative pairs. $D$ means the document. Finally, our model is trained by minimizing the infoNCE \cite{radford2021learning} loss.

\textbf{Model Inference.} Given a image/caption pair as the query $x^{ic}_{k,i}$, the cosine similarity between the query and all sentences in the document is computed. Then we sort the sentences in descending order according to the similarity and the first candidate will be picked as the most relevant sentence prediction. We also regard the section including this sentence as the most relevant section prediction.

\label{sec:model}

\section{Pretraining Dataset}
\label{sec:dataset}
In this section, we describe the data creation procedure of our pretraining dataset: Wikipedia dataset and present dataset statistic. First, we used the latest English Wikipedia dump\footnote{https://github.com/muraoka7/tool4ipp} to extract Wikipedia articles \cite{muraoka2020image}. In Wikipedia articles, images are nested in two HTML tags, which delineate the sections. We regard all the sentences within the same section as the ground truth since obtaining sentence-level associations would require additional manual labeling. \cite{tran2020transform}'s dataset has the groundtruth linking between image/caption with sentences in the news articles. However, we don't have the right to use them according to the official website\footnote{https://help.nytimes.com/hc/en-us/articles/115014893428-Terms-of-Service\#2}.

\begin{figure}[t]
    \centering
      \includegraphics[width=0.46\textwidth]{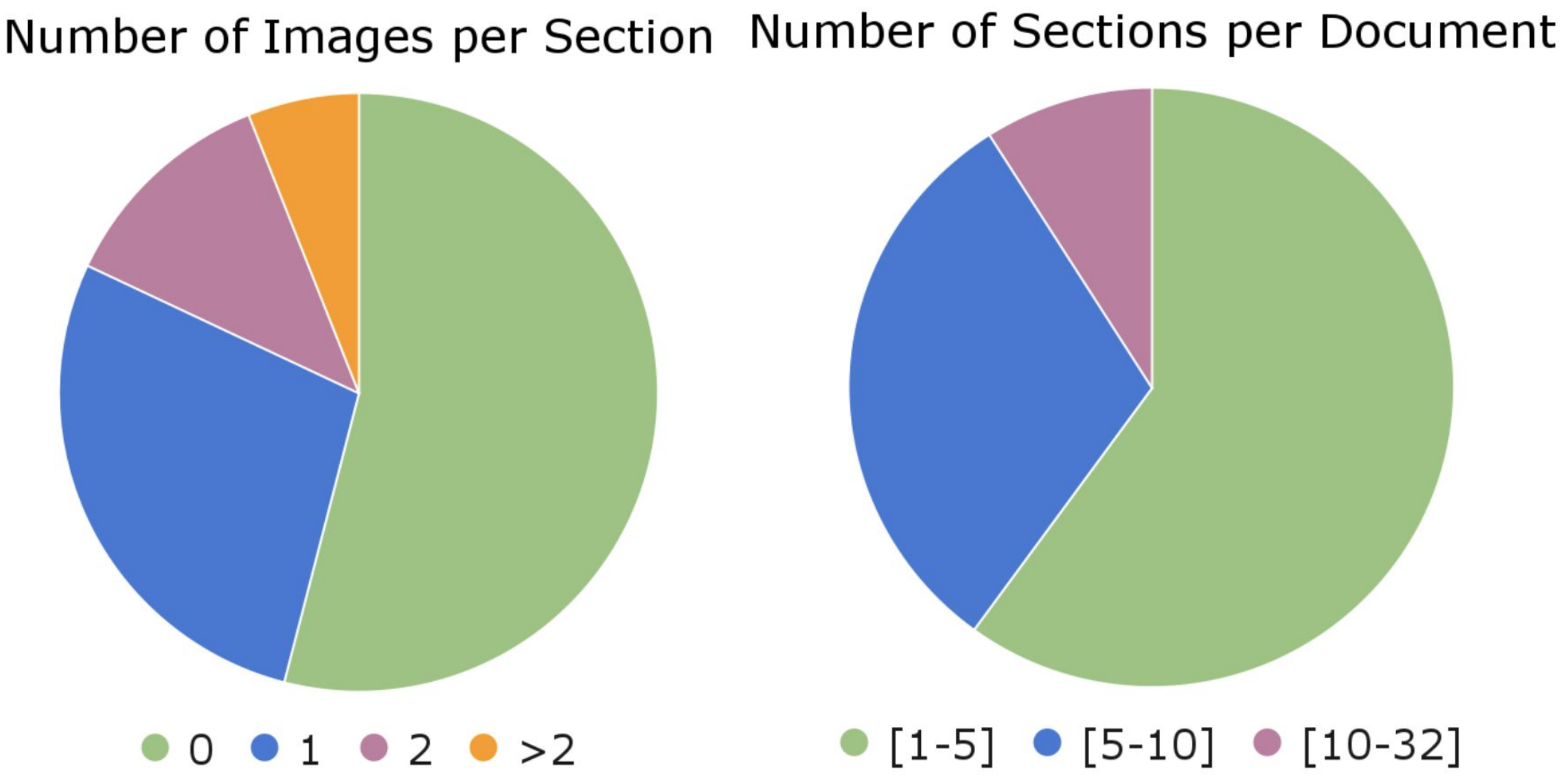}
    \vspace{-2pt}
    \caption{Distributions of the image number per section and section number per document in our Wikipedia dataset.
    }
    \label{fig:dataset}
\end{figure}

Consequently, we obtained 150,656 documents, not including ones failed to be parsed. Then we only kept documents with 2 to 30 images and no more than 32 sections. We also converted all the extracted images to an RGB format in order to guarantee quality. Eventually, this resulted in 66k articles and 320k images. We summarize the statistics of our dataset in Table~\ref{tab:dataset_summary} with previous datasets \cite{shah2019kvqa,kembhavi2017you,liu2020visual,furkan2019good}. Each section in our dataset has 8.14 sentences on average, the average sentence length is 22.3 words and the average section length is 195.5 words in our dataset, which is the largest among the compared datasets. In addition, over 50\% documents in our dataset have more than 10 sections and many sections don’t have corresponding images (Figure~\ref{fig:dataset}). Compared to other datasets used in Image-Text alignment methods (e.g. COCO captioning), the amount of text associated with an image is much larger. This confirms that understanding of longer texts is the required ability.

. 

\section{Experiments}
\label{sec:experiment}
In this section, we first introduce details of our implementation and present comprehensive experiment results.
\begin{table}[t]
\setlength\tabcolsep{1.5pt}
\centering
\small
\begin{tabular}{lccc}
\toprule
Model & R@1 & R@3 & A-R@1\\
\midrule
CLIP(\textit{ViT32}) \cite{radford2021learning} &  0.40 & 0.69 & 0.39\\
CLIP(\textit{RN50}) \cite{radford2021learning} &  0.39 & 0.67 & 0.38\\
SBert \cite{reimers2019sentence} &  0.38 & 0.67 & 0.38\\
Pythia \cite{muraoka2020image}  & 0.33 & 0.55 & 0.34\\
Newsroom \cite{liu2020upgrading}  & 0.31 & 0.51 & 0.31\\
SANDI \cite{nag2021sandi}  & 0.34 & 0.57 & 0.33\\
LAIS \cite{zhang2021news}  & 0.45 & 0.74 & 0.45\\
\midrule
\textbf{\textit{DocumentCLIP}}\textit{-Early Fusion}  & 0.57 & 0.87 & 0.56\\
\textbf{\textit{DocumentCLIP}}\textit{-Late Fusion}  & 0.56 & 0.85 & 0.54\\
\midrule
\textbf{\textit{DocumentCLIP}}\textit{-Early Fusion}  & 0.57 & 0.87 & 0.56\\
-w/o salience-aware loss  & 0.54 & 0.84 & 0.53\\
\quad-w/o entity check  & 0.46 & 0.76 & 0.46\\
\quad\quad-w/o layout information  & 0.39 & 0.67 & 0.39\\
\bottomrule
\end{tabular}
\caption{Comparative Experiments with baselines and ablation study on Wikipedia datasets.
}
\vspace{-0.2in}
\label{tab:baselines}
\end{table}

\vspace{-0.15in}
\subsection{Experimental Details}
\textbf{Evaluation Setting.} \underline{Supervised settings}: We first evaluate the most relevant section prediction on the Wikipedia dataset in both the image level and document level.  At the image level, we use \textit{R@N(N=1,3)} to calculate the percentage of images where the groundtruth section is among the \textit{TopN} predicted sections. At the document level, we measure the percentage of documents where all images in the document have a correct \textit{Top1} predicted section: $\textit{A-R@1}$. \underline{Zero-shot setting}: We ask human annotators to evaluate the most relevant sentence prediction on the \textit{Open Textbook} dataset. \textit{DocumentCLIP} is trained on the Wikipedia dataset and directly tested in the zero-shot setting. More detailed more described in \ref{human}.

\textbf{Implementation Details.} As for the experiment setting, we randomly split our dataset into 80\% for training, 10\% for validation and 10\% for testing. We trained our model with mini-batches of size 32 by using stochastic gradient descent. We used AdamW \cite{you2019large} as an optimization algorithm and a warm-up strategy with the initial learning rate of 0.001.

\textbf{Baseline Models. }We run the public release of \textit{CLIP} \cite{radford2021learning} and \textit{SBert} \cite{reimers2019sentence} and finetune them on the Wikipedia dataset in the supervised task. As for the zero-shot setting, they aren't finedtuned. \textit{CLIP} takes both images and captions as the input while \textit{SBert} only takes the captions. Additionally, only the first sentence in each section is fed into \textit{CLIP} since its maximum length of text encoder is 77, which is much below the average section length. \textit{Pythia} \cite{muraoka2020image} encodes different modalities separately and computes an attention layer to attend specific parts in the image. \textit{Newsroom} \cite{muraoka2020image} adopts a hierarchical self-attention mechanism to attend to both key words within a piece of captions and informative components of a document. \textit{LAIS} \cite{zhang2021news} introduce a two-stage architecture to select the insertion position for the images.

\begin{table}[t]
\setlength\tabcolsep{2pt}
\centering
\small
\begin{tabular}{lccc}
\toprule
\quad \; Training &  Inference & R@1 & R@3\\
\textit{Image}\quad \textit{Caption} &  \textit{Image}\quad \textit{Caption} &  & \\
\midrule
\cmark \quad\quad\quad \textit{Original} &  \quad\quad\quad\cmark \quad\quad \textit{Original} & 0.57 & 0.87\\
\cmark \quad\quad\quad \textit{Generated} &  \quad\quad\quad\quad\cmark \quad\quad \textit{Generated} & 0.51 & 0.82\\
\xmark \quad\quad\quad \textit{Original} &  \quad\quad\quad \xmark \quad\quad \textit{Original} & 0.54 & 0.84\\
\xmark \quad\quad\quad \textit{Generated} &  \quad\quad\quad\quad \xmark \quad\quad \textit{Generated} & 0.49 & 0.79\\
\cmark \quad\quad\quad \xmark &   \cmark \quad\quad \xmark & 0.50 & 0.81\\
\midrule
\cmark \quad\quad\quad \textit{Original} &  \cmark \quad\quad \textit{\xmark} & 0.52 & 0.82\\
\cmark \quad\quad\quad \textit{Original} &  \quad\quad\quad \xmark \quad\quad \textit{Original} & 0.55 & 0.85\\
\bottomrule
\end{tabular}
\caption{Performance of $\textit{DocumentCLIP}$ with or without images and captions. $\textit{Generated}$ captions are from \textit{BLIP} \cite{li2022blip}.}
\vspace{-0.2in}
\label{tab:modalities}
\end{table}

\begin{table}[t]
\setlength\tabcolsep{4.5pt}
\centering
\small
\begin{tabular}{lccc}
\toprule
Section Encoding &  Initialization & R@1 & R@3\\
\midrule
\textit{First Sentence} &  \textit{CLIP(ViT32)} & 0.54 & 0.84\\
\textit{Weighted Average} &  \textit{CLIP(ViT32)}& 0.55 & 0.85\\
\textit{All Sentences Concat} & \textit{CLIP(ViT32)}&  0.52 & 0.82\\
\midrule
\textit{Salient Sentence} &  \textit{CLIP(ViT32)} & 0.57 & 0.87\\
\textit{Salient Sentence} &  \textit{CLIP(RN50)} & 0.55 & 0.86\\
\textit{Salient Sentence} &  \textit{ROBERTA*} & 0.56 & 0.86\\
\textit{Salient Sentence} &  \textit{SCRATCH} & 0.49 & 0.79\\
\bottomrule
\end{tabular}
\caption{Performance of $\textit{DocumentCLIP}$ to investigate different section encoding strategies and initialization methods. $\textit{ROBERTA*}$ means using the vision transformer $\textit{CLIP(ViT32)}$ to encode images and $\textit{Roberta}$ \cite{liu2019roberta} to encode text.
}
\vspace{-0.2in}
\label{tab:section_encoding}
\end{table}

\begin{table}[t]
\setlength\tabcolsep{4.5pt}
\centering
\small
\begin{tabular}{lccc}
\toprule
Section Encoding &  Top$K$ & R@1 & R@3\\
\midrule
\textit{Salient Sentence} &  $K=1$ & 0.54 & 0.84\\
\textit{Salient Sentence} &  $K=3$ & 0.56 & 0.86\\
\textit{Salient Sentence} &  $K=5$ & 0.57 & 0.87\\
\textit{Salient Sentence} &  $K=7$ & 0.57 & 0.86\\
\textit{Salient Sentence} &  $K=9$ & 0.55 & 0.84\\
\bottomrule
\end{tabular}
\caption{Performance of  different number of candidates (Top $K$) in the salient sentences extraction. }
\vspace{-0.2in}
\label{tab:topk}
\end{table}

\subsection{Supervised Experiment Results}
\subsubsection{Comparison with Baselines}
We report the overall results of most relevant section prediction in the supervised setting on the Wikipedia dataset (Table~\ref{tab:baselines}). We make the following observations:
our model $\textit{DocumentCLIP}$ achieves more competitive performance in all scores by a large margin up to 10\%. This remarkable improvement indicates that $\textit{DocumentCLIP}$ successfully possesses the intra-document alignment between image and text. The major reason is that previous methods aren't able to distinguish between the sections in the same document since these sections have similar topics. Instead, we use salience-aware contrastive learning with layout features to learn the document context and compare common entities with captions to differentiate between sections. 

\begin{figure}[t]
    \centering
      \includegraphics[width=0.48\textwidth]{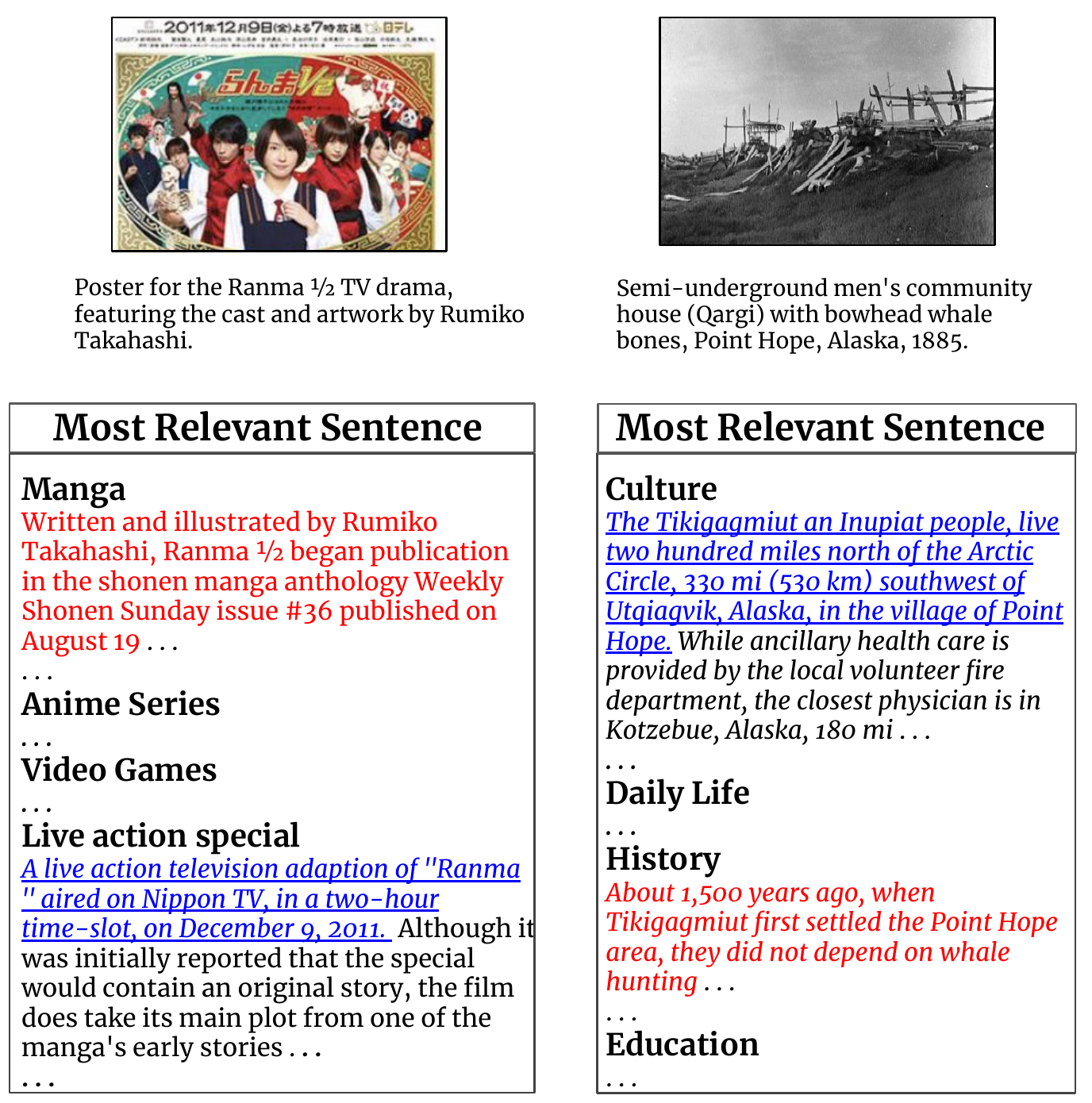}
    \caption{Two examples of Wikipedia articles. The red section is the prediction from baseline method while the blue section is the prediction by \textit{DocumentCLIP}. The sentence in blue is the most relevant candidate.
    }
    \vspace{-0.2in}s
    \label{fig:example_from_wiki}
\end{figure}

\subsubsection{Ablation Studies}
\textbf{Contributions of Different Components.} The below sections of Table~\ref{tab:baselines} summarizes the ablative performance of our model with different components. Without layout information, the performance shows 7\% degradation in terms of $\textit{R@1}$ and 9\% degradation in $\textit{R@3}$, which demonstrates that the structural information like the position features and segment features are essential for our model to learn the correlation and difference between each sections and how to connect to the semantic of image/caption pairs. We notice that \textit{DocumentCLIP} benefits from entity check and salient-aware loss since they can not only filter out noisy sentences in sections but also check keyword alignment between captions and sections. Additionally, the improvement from early fusion strategy indicates the its efficiency to aggregate multimodal features between the image and caption.

\begin{table}[t]
\setlength\tabcolsep{4.7pt}
\centering
\small
\begin{tabular}{lc|ccc}
\toprule
Model &  Avg Rank  & Rank \#1 & Rank \#2 & Rank \#3\\
\midrule
\textit{SBert} &  1.29 & 22\% & 31\% & 47\%\\
\textit{CLIP} &  1.13 & 29\% & 36\% & 35\%\\
\textbf{\textit{DocumentCLIP}} &  \textbf{0.75} & \textbf{49\%} & 33\% & 18\%\\
\bottomrule
\end{tabular}
\caption{Human evaluation of different models on \textit{Open Textbook} dataset in the zero-shot setting. Annotators are asked to put the sentences in order of relevance to the Image/Caption pair. "\textit{1.29}" means the average rank of \textit{SBert} is 1.29. "\textit{22\%}" means only 22\% of the predictions from \textit{SBert} are put in the 1st place. }
\vspace{-0.2in}
\label{tab:human}
\end{table}

\textbf{Importance of Different Modalities.} Besides, we investigate the contribution of each modality (image or caption) in Table~\ref{tab:modalities}. Given only one modality in training and inference, the performance of captions is superior to that of images (third and fifth row). This can be reasonable because captions sometimes share similar or same words with textual sections. To further analyze the effect of captions, we replace the original captions with the ones generated by $\textit{BLIP}$ \cite{li2022blip}, an state-of-art and pretrained vision-language generation model (second and forth row). The performance degradation is expected since generated captions always miss the key entities in the original ones, which play the evidence role. 

\begin{figure*}[t!]
    \centering
      \includegraphics[width=0.80\textwidth]{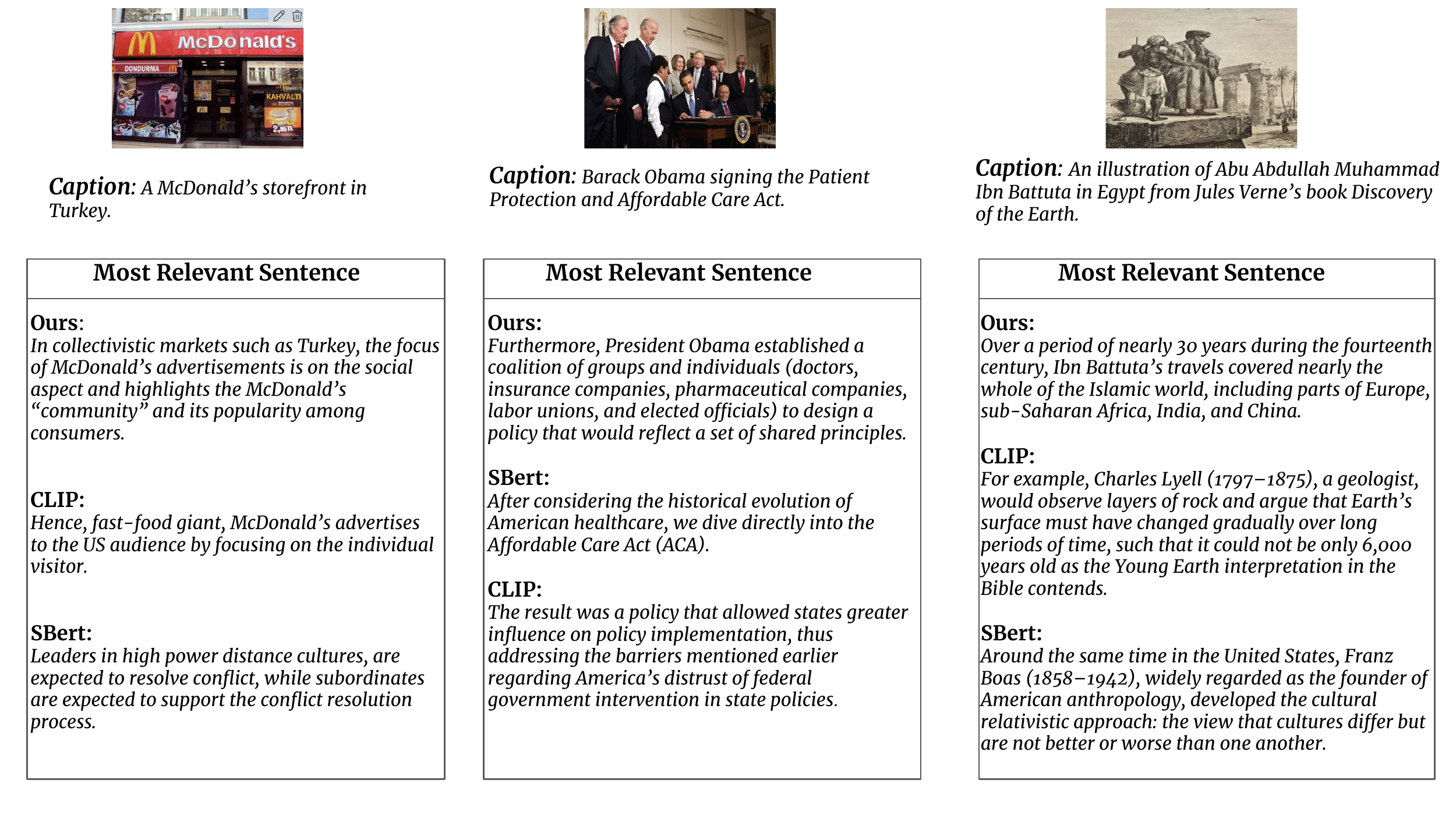}
    \caption{Selected image/caption samples and most relevant sentence predictions from Open Textbook dataset. We compare our predictions with the ones from \textit{CLIP}, \textit{SBert} and demonstrate the advance of \textit{DocumentCLIP}. }
    \vspace{-0.2in}
    \label{fig:vs1}
\end{figure*}

\textbf{Section Encoding Strategies.} Table~\ref{tab:section_encoding} studies the effect of different section encoding strategies. Compared to other strategies, the improvement is evident when using salient sentence extraction. The main reason for this might because other strategies fail to remove the influence of the noisy sentences, which will mislead the model to pick the wrong the section. To further investigate its effect, we gradually increase the Top$K$ value and implement the salient sentence extraction in Table~\ref{tab:topk}. We notice that a larger $K$ will yield better accuracy when $K$ is smaller than 5. The phenomenon indicates that single sentence isn't informational enough to represent the section while too many ones will introduce additional noise.

\textbf{Initialization.} We also analyze different initialization methods in Table~\ref{tab:section_encoding}. As the $\textit{ROBERTA*}$ (six row) method suggests, we use the vision transformer $\textit{CLIP(ViT32)}$ to encode images and $\textit{Roberta}$ \cite{liu2019roberta} to encode text. We found it achieves best results if we use $\textit{CLIP}$ to encode both images and captions. This is because $\textit{CLIP}$ is trained on a large vision and language dataset, alleviating the semantic gap between images and text. Besides, $\textit{SCRATCH}$ performs worst due to missing the prior knowledge.

\vspace{-0.1in}
\subsubsection{Qualitative Analysis. }In order to evaluate the quality of the complementary text retrieved by our model in the sentence level, examples are shown in Figure~\ref{fig:example_from_wiki}. This left document talks about a Japanese manga series called Ranma ½. The image/caption pair is in the 2nd place and talks about the TV drama version of Ranma ½. Previous methods regard the first section as the prediction since it share more common entities with the caption. However, they fail to encode image/pair position information and understand commonsense knowledge. For example, $\textit{"live action special"}$ from the fifth section actually refers to $\textit{"TV drama"}$ in the caption. As for the right document, current methods only consider the keywords within the captions without the image position information.

\subsection{Zero-Shot Experiment Results}\label{human}
\vspace{-0.05in}
In this section, we gain further insights into the most relevant sentence prediction on the \textit{Open Textbook} dataset via human evaluation. With the human evaluation, we access whether our model could pick more relevant sentences than \textit{CLIP} \cite{radford2021learning} and \textit{SBert} \cite{reimers2019sentence} in the zero-shot setting. 

Here, we randomly select a set of 40 image/captions pairs from 10 articles in \textit{Open Textbook} dataset. We conduct our evaluations on 75 workers from Amazon Mechanical Turk\footnote{https://www.mturk.com}. The 40 questions in the questionnaire are shuffled for each worker and it takes 40 minutes to complete on average. We also add 3 attention questions to guarantee the quality of the response. As for each question, they are given a caption and image pair and asked to put the three sentences in order of relevance to the Image/Caption pair and the most relevant should be in the 1st place. Summary of the evaluation are in Tab~\ref{tab:human}. The average rank of the predictions from \textit{DocumentCLIP} is 0.75, which is much lower than \textit{SBert}: 1.29 and \textit{CLIP}: 1.13. In addition, 49\% of the samples have the \textit{DocumentCLIP} in the 1st place and only 18\% of the samples have the \textit{DocumentCLIP} in the final place. Figure~\ref{fig:vs1} demonstrate the advance of our model to understand the intra-document multimodal interaction.

\vspace{-0.1in}

\section{Related Work}
\vspace{-0.1in}
Cross-modal research \cite{li2022blip, alayrac2022flamingo, li2023blip, liu2023covid, zhang2021vinvl, yao2021filip, xu2017video, liu2023covid, Liu2020VisualNB, liu2020visual, Liu2020VisualNewsA, Liu2020VisualNewsB, Liu2023AligningLM, Li2023TowardsUI} has recently received increasing attention as a result of the advances in both computer vision and natural language processing. \cite{huang2017instance, lee2018stacked, zhang2018deep, wang2019matching} encode multimodal features into a common embedding space in which instance similarity is measured by conventional cosine or Euclidean distance. Recent state-of-the-art works \cite{radford2021learning, li2019visualbert, lu2019vilbert, su2019vl} employ pretraining methods, cross transformers and contrastive learning algorithms to learn the correlation between different modalities. Various pre-training objectives have also been proposed over the years, and have progressively converged to a few time-tested ones \cite{yao2021filip, li2021align, li2020oscar, yu2022coca, zellers2021merlot, zellers2022merlot, zhai2022lit}. However, the limitation is that they are trained to align short literal text descriptions with images but fail to handle real-world media such as news articles \cite{liu2020visual}, Wikipedia pages \cite{muraoka2020image}, magazines, product descriptions consisting of multiple sentences with multiple images and layout information. Document structure analysis models \cite{huang2022layoutlmv3, gu2021unidoc, xu2020layoutlmv2, appalaraju2021docformer} process pages one by one, but the figures may relate to text located on other pages. 

Different from existing methods, \textit{DocumentCLIP} is not only able to effectively process long text, but also understand text-image interaction at the sentence-level in the document.
\label{sec:related_work}

\vspace{-0.15in}
\section{Conclusions}

In this paper, we are the first to explore the sentence-level multimodal intra-document links in vision-language pretraining. We also introduce a novel contrastive learning framework with layout information, multimodal interaction, novel section encoding strategy, salience-aware contrastive loss and hard negative samples. We also collect a large visual document dataset from Wikipedia with 66k articles, 320k images/caption pairs and various topics. Furthermore, our proposed model \textit{DocumentCLIP} achieves state-of-the-art performance in both the supervised and zero-shot settings. We hope this work paves the way for future studies in document understanding and multi-modal alignment.
\label{sec:conclusion}

{\small
\bibliographystyle{ieee_fullname}
\bibliography{egbib}

\begin{thebibliography}{10}\itemsep=-1pt

\bibitem{alayrac2022flamingo}
Jean-Baptiste Alayrac, Jeff Donahue, Pauline Luc, Antoine Miech, Iain Barr,
  Yana Hasson, Karel Lenc, Arthur Mensch, Katie Millican, Malcolm Reynolds,
  et~al.
\newblock Flamingo: a visual language model for few-shot learning.
\newblock {\em arXiv preprint arXiv:2204.14198}, 2022.

\bibitem{appalaraju2021docformer}
Srikar Appalaraju, Bhavan Jasani, Bhargava~Urala Kota, Yusheng Xie, and R
  Manmatha.
\newblock Docformer: End-to-end transformer for document understanding.
\newblock In {\em Proceedings of the IEEE/CVF international conference on
  computer vision}, pages 993--1003, 2021.

\bibitem{furkan2019good}
Ali Furkan~Biten, Lluis Gomez, Mar{\c{c}}al Rusi{\~n}ol, and Dimosthenis
  Karatzas.
\newblock Good news, everyone! context driven entity-aware captioning for news
  images.
\newblock {\em arXiv e-prints}, pages arXiv--1904, 2019.

\bibitem{gu2021unidoc}
Jiuxiang Gu, Jason Kuen, Vlad~I Morariu, Handong Zhao, Rajiv Jain, Nikolaos
  Barmpalios, Ani Nenkova, and Tong Sun.
\newblock Unidoc: Unified pretraining framework for document understanding.
\newblock {\em Advances in Neural Information Processing Systems}, 34:39--50,
  2021.

\bibitem{huang2022layoutlmv3}
Yupan Huang, Tengchao Lv, Lei Cui, Yutong Lu, and Furu Wei.
\newblock Layoutlmv3: Pre-training for document ai with unified text and image
  masking.
\newblock {\em arXiv preprint arXiv:2204.08387}, 2022.

\bibitem{huang2017instance}
Yan Huang, Wei Wang, and Liang Wang.
\newblock Instance-aware image and sentence matching with selective multimodal
  lstm.
\newblock In {\em Proceedings of the IEEE Conference on Computer Vision and
  Pattern Recognition}, pages 2310--2318, 2017.

\bibitem{kembhavi2017you}
Aniruddha Kembhavi, Minjoon Seo, Dustin Schwenk, Jonghyun Choi, Ali Farhadi,
  and Hannaneh Hajishirzi.
\newblock Are you smarter than a sixth grader? textbook question answering for
  multimodal machine comprehension.
\newblock In {\em Proceedings of the IEEE Conference on Computer Vision and
  Pattern recognition}, pages 4999--5007, 2017.

\bibitem{lee2018stacked}
Kuang-Huei Lee, Xi Chen, Gang Hua, Houdong Hu, and Xiaodong He.
\newblock Stacked cross attention for image-text matching.
\newblock In {\em Proceedings of the European conference on computer vision
  (ECCV)}, pages 201--216, 2018.

\bibitem{li2023blip}
Junnan Li, Dongxu Li, Silvio Savarese, and Steven Hoi.
\newblock Blip-2: Bootstrapping language-image pre-training with frozen image
  encoders and large language models.
\newblock {\em arXiv preprint arXiv:2301.12597}, 2023.

\bibitem{li2022blip}
Junnan Li, Dongxu Li, Caiming Xiong, and Steven Hoi.
\newblock Blip: Bootstrapping language-image pre-training for unified
  vision-language understanding and generation.
\newblock {\em arXiv preprint arXiv:2201.12086}, 2022.

\bibitem{li2021align}
Junnan Li, Ramprasaath Selvaraju, Akhilesh Gotmare, Shafiq Joty, Caiming Xiong,
  and Steven Chu~Hong Hoi.
\newblock Align before fuse: Vision and language representation learning with
  momentum distillation.
\newblock {\em Advances in neural information processing systems},
  34:9694--9705, 2021.

\bibitem{li2020hero}
Linjie Li, Yen-Chun Chen, Yu Cheng, Zhe Gan, Licheng Yu, and Jingjing Liu.
\newblock Hero: Hierarchical encoder for video+ language omni-representation
  pre-training.
\newblock {\em arXiv preprint arXiv:2005.00200}, 2020.

\bibitem{li2019visualbert}
Liunian~Harold Li, Mark Yatskar, Da Yin, Cho-Jui Hsieh, and Kai-Wei Chang.
\newblock Visualbert: A simple and performant baseline for vision and language.
\newblock {\em arXiv preprint arXiv:1908.03557}, 2019.

\bibitem{li2020oscar}
Xiujun Li, Xi Yin, Chunyuan Li, Pengchuan Zhang, Xiaowei Hu, Lei Zhang, Lijuan
  Wang, Houdong Hu, Li Dong, Furu Wei, et~al.
\newblock Oscar: Object-semantics aligned pre-training for vision-language
  tasks.
\newblock In {\em Computer Vision--ECCV 2020: 16th European Conference,
  Glasgow, UK, August 23--28, 2020, Proceedings, Part XXX 16}, pages 121--137.
  Springer, 2020.

\bibitem{Li2023TowardsUI}
Zongxi Li, Paiheng Xu, Fuxiao Liu, and Hyemi Song.
\newblock Towards understanding in-context learning with contrastive
  demonstrations and saliency maps.
\newblock {\em ArXiv}, abs/2307.05052, 2023.

\bibitem{lin2014microsoft}
Tsung-Yi Lin, Michael Maire, Serge Belongie, James Hays, Pietro Perona, Deva
  Ramanan, Piotr Doll{\'a}r, and C~Lawrence Zitnick.
\newblock Microsoft coco: Common objects in context.
\newblock In {\em European conference on computer vision}, pages 740--755.
  Springer, 2014.

\bibitem{liu2020upgrading}
Fangyu Liu, R{\'e}mi Lebret, Didier Orel, Philippe Sordet, and Karl Aberer.
\newblock Upgrading the newsroom: an automated image selection system for news
  articles.
\newblock {\em ACM Transactions on Multimedia Computing, Communications, and
  Applications (TOMM)}, 16(3):1--28, 2020.

\bibitem{Liu2023AligningLM}
Fuxiao Liu, Kevin Lin, Linjie Li, Jianfeng Wang, Yaser Yacoob, and Lijuan Wang.
\newblock Aligning large multi-modal model with robust instruction tuning.
\newblock {\em ArXiv}, abs/2306.14565, 2023.

\bibitem{liu2020visual}
Fuxiao Liu, Yinghan Wang, Tianlu Wang, and Vicente Ordonez.
\newblock Visual news: Benchmark and challenges in news image captioning.
\newblock {\em arXiv preprint arXiv:2010.03743}, 2020.

\bibitem{Liu2020VisualNB}
Fuxiao Liu, Yinghan Wang, Tianlu Wang, and Vicente Ordonez.
\newblock Visual news: Benchmark and challenges in news image captioning.
\newblock In {\em Conference on Empirical Methods in Natural Language
  Processing}, 2020.

\bibitem{Liu2020VisualNewsA}
Fuxiao Liu, Yinghan Wang, Tianlu Wang, and Vicente Ordonez.
\newblock Visualnews : A large multi-source news image dataset.
\newblock {\em arXiv: Computer Vision and Pattern Recognition}, 2020.

\bibitem{Liu2020VisualNewsB}
Fuxiao Liu, Yinghan Wang, Tianlu Wang, and Vicente Ordonez.
\newblock Visualnews : Benchmark and challenges in entity-aware image
  captioning.
\newblock {\em ArXiv}, abs/2010.03743, 2020.

\bibitem{liu2023covid}
Fuxiao Liu, Yaser Yacoob, and Abhinav Shrivastava.
\newblock Covid-vts: Fact extraction and verification on short video platforms.
\newblock {\em arXiv preprint arXiv:2302.07919}, 2023.

\bibitem{liu2019roberta}
Yinhan Liu, Myle Ott, Naman Goyal, Jingfei Du, Mandar Joshi, Danqi Chen, Omer
  Levy, Mike Lewis, Luke Zettlemoyer, and Veselin Stoyanov.
\newblock Roberta: A robustly optimized bert pretraining approach.
\newblock {\em arXiv preprint arXiv:1907.11692}, 2019.

\bibitem{lu2019vilbert}
Jiasen Lu, Dhruv Batra, Devi Parikh, and Stefan Lee.
\newblock Vilbert: Pretraining task-agnostic visiolinguistic representations
  for vision-and-language tasks.
\newblock {\em Advances in neural information processing systems}, 32, 2019.

\bibitem{muraoka2020image}
Masayasu Muraoka, Ryosuke Kohita, and Etsuko Ishii.
\newblock Image position prediction in multimodal documents.
\newblock In {\em Proceedings of the Twelfth Language Resources and Evaluation
  Conference}, pages 4265--4274, 2020.

\bibitem{nag2021sandi}
Sreyasi Nag~Chowdhury, Simon Razniewski, and Gerhard Weikum.
\newblock Sandi: Story-and-images alignment.
\newblock In {\em 16th Conference of the European Chapter of the Association
  for Computational Linguistics}, pages 989--999. ACL, 2021.

\bibitem{radford2021learning}
Alec Radford, Jong~Wook Kim, Chris Hallacy, Aditya Ramesh, Gabriel Goh,
  Sandhini Agarwal, Girish Sastry, Amanda Askell, Pamela Mishkin, Jack Clark,
  et~al.
\newblock Learning transferable visual models from natural language
  supervision.
\newblock In {\em International Conference on Machine Learning}, pages
  8748--8763. PMLR, 2021.

\bibitem{radford2019language}
Alec Radford, Jeffrey Wu, Rewon Child, David Luan, Dario Amodei, Ilya
  Sutskever, et~al.
\newblock Language models are unsupervised multitask learners.
\newblock {\em OpenAI blog}, 1(8):9, 2019.

\bibitem{reimers2019sentence}
Nils Reimers and Iryna Gurevych.
\newblock Sentence-bert: Sentence embeddings using siamese bert-networks.
\newblock {\em arXiv preprint arXiv:1908.10084}, 2019.

\bibitem{shah2019kvqa}
Sanket Shah, Anand Mishra, Naganand Yadati, and Partha~Pratim Talukdar.
\newblock Kvqa: Knowledge-aware visual question answering.
\newblock In {\em Proceedings of the AAAI conference on artificial
  intelligence}, volume~33, pages 8876--8884, 2019.

\bibitem{shibata1999byte}
Yusuxke Shibata, Takuya Kida, Shuichi Fukamachi, Masayuki Takeda, Ayumi
  Shinohara, Takeshi Shinohara, and Setsuo Arikawa.
\newblock Byte pair encoding: A text compression scheme that accelerates
  pattern matching.
\newblock 1999.

\bibitem{su2019vl}
Weijie Su, Xizhou Zhu, Yue Cao, Bin Li, Lewei Lu, Furu Wei, and Jifeng Dai.
\newblock Vl-bert: Pre-training of generic visual-linguistic representations.
\newblock {\em arXiv preprint arXiv:1908.08530}, 2019.

\bibitem{tran2020transform}
Alasdair Tran, Alexander Mathews, and Lexing Xie.
\newblock Transform and tell: Entity-aware news image captioning.
\newblock In {\em Proceedings of the IEEE/CVF Conference on Computer Vision and
  Pattern Recognition}, pages 13035--13045, 2020.

\bibitem{wang2019matching}
Tan Wang, Xing Xu, Yang Yang, Alan Hanjalic, Heng~Tao Shen, and Jingkuan Song.
\newblock Matching images and text with multi-modal tensor fusion and
  re-ranking.
\newblock In {\em Proceedings of the 27th ACM international conference on
  multimedia}, pages 12--20, 2019.

\bibitem{xu2017video}
Dejing Xu, Zhou Zhao, Jun Xiao, Fei Wu, Hanwang Zhang, Xiangnan He, and Yueting
  Zhuang.
\newblock Video question answering via gradually refined attention over
  appearance and motion.
\newblock In {\em Proceedings of the 25th ACM international conference on
  Multimedia}, pages 1645--1653, 2017.

\bibitem{xu2020layoutlmv2}
Yang Xu, Yiheng Xu, Tengchao Lv, Lei Cui, Furu Wei, Guoxin Wang, Yijuan Lu,
  Dinei Florencio, Cha Zhang, Wanxiang Che, et~al.
\newblock Layoutlmv2: Multi-modal pre-training for visually-rich document
  understanding.
\newblock {\em arXiv preprint arXiv:2012.14740}, 2020.

\bibitem{yao2021filip}
Lewei Yao, Runhui Huang, Lu Hou, Guansong Lu, Minzhe Niu, Hang Xu, Xiaodan
  Liang, Zhenguo Li, Xin Jiang, and Chunjing Xu.
\newblock Filip: fine-grained interactive language-image pre-training.
\newblock {\em arXiv preprint arXiv:2111.07783}, 2021.

\bibitem{you2019large}
Yang You, Jing Li, Sashank Reddi, Jonathan Hseu, Sanjiv Kumar, Srinadh
  Bhojanapalli, Xiaodan Song, James Demmel, Kurt Keutzer, and Cho-Jui Hsieh.
\newblock Large batch optimization for deep learning: Training bert in 76
  minutes.
\newblock {\em arXiv preprint arXiv:1904.00962}, 2019.

\bibitem{yu2022coca}
Jiahui Yu, Zirui Wang, Vijay Vasudevan, Legg Yeung, Mojtaba Seyedhosseini, and
  Yonghui Wu.
\newblock Coca: Contrastive captioners are image-text foundation models.
\newblock {\em arXiv preprint arXiv:2205.01917}, 2022.

\bibitem{zellers2022merlot}
Rowan Zellers, Jiasen Lu, Ximing Lu, Youngjae Yu, Yanpeng Zhao, Mohammadreza
  Salehi, Aditya Kusupati, Jack Hessel, Ali Farhadi, and Yejin Choi.
\newblock Merlot reserve: Neural script knowledge through vision and language
  and sound.
\newblock In {\em Proceedings of the IEEE/CVF Conference on Computer Vision and
  Pattern Recognition}, pages 16375--16387, 2022.

\bibitem{zellers2021merlot}
Rowan Zellers, Ximing Lu, Jack Hessel, Youngjae Yu, Jae~Sung Park, Jize Cao,
  Ali Farhadi, and Yejin Choi.
\newblock Merlot: Multimodal neural script knowledge models.
\newblock {\em Advances in Neural Information Processing Systems},
  34:23634--23651, 2021.

\bibitem{zhai2022lit}
Xiaohua Zhai, Xiao Wang, Basil Mustafa, Andreas Steiner, Daniel Keysers,
  Alexander Kolesnikov, and Lucas Beyer.
\newblock Lit: Zero-shot transfer with locked-image text tuning.
\newblock In {\em Proceedings of the IEEE/CVF Conference on Computer Vision and
  Pattern Recognition}, pages 18123--18133, 2022.

\bibitem{zhang2021vinvl}
Pengchuan Zhang, Xiujun Li, Xiaowei Hu, Jianwei Yang, Lei Zhang, Lijuan Wang,
  Yejin Choi, and Jianfeng Gao.
\newblock Vinvl: Making visual representations matter in vision-language
  models.
\newblock {\em arXiv preprint arXiv:2101.00529}, 1(6):8, 2021.

\bibitem{zhang2018deep}
Ying Zhang and Huchuan Lu.
\newblock Deep cross-modal projection learning for image-text matching.
\newblock In {\em Proceedings of the European conference on computer vision
  (ECCV)}, pages 686--701, 2018.

\bibitem{zhang2021news}
Zhengkun Zhang, Jun Wang, Adam Jatowt, Zhe Sun, Shao-Ping Lu, and Zhenglu Yang.
\newblock News content completion with location-aware image selection.
\newblock In {\em Proceedings of the AAAI Conference on Artificial
  Intelligence}, volume~35, pages 14498--14505, 2021.

\end{thebibliography}
}

\end{document}